# Automated Scoring of Graphical Open-Ended Responses Using Artificial Neural Networks

Matthias von Davier[1], Lillian Tyack, Lale Khorramdel

*TIMSS & PIRLS International Study Center at Boston College*

---

[1] Boston College, 188-194 Beacon Street, Chestnut Hill, MA 02467, USA, email: vondavim@bc.edu

Abstract

Automated scoring of free drawings or images as responses has yet to be utilized in large-scale assessments of student achievement. In this study, we propose artificial neural networks to classify these types of graphical responses from a computer based international mathematics and science assessment. We are comparing classification accuracy of convolutional and feedforward approaches. Our results show that convolutional neural networks (CNNs) outperform feedforward neural networks in both loss and accuracy. The CNN models classified up to 97.71% of the image responses into the appropriate scoring category, which is comparable to, if not more accurate, than typical human raters. These findings were further strengthened by the observation that the most accurate CNN models correctly classified some image responses that had been incorrectly scored by the human raters. As an additional innovation, we outline a method to select human rated responses for the training sample based on an application of the expected response function derived from item response theory. This paper argues that CNN-based automated scoring of image responses is a highly accurate procedure that could potentially replace the workload and cost of second human raters for large scale assessments, while improving the validity and comparability of scoring complex constructed-response items.

Introduction

The transition to a computer-based assessment of student skills has enabled international large-scale assessments (ILSAs) to improve measurement through innovative item-types including constructed-response items. Not only are constructed-response items considered more engaging than traditional item types (e.g., multiple-choice items), but they also capture complex constructs with greater nuance and precision (Bejar, 2017; Zhai, 2021). The TIMSS 2019 digital assessment, for example, incorporated many constructed-response items, including grid-based image responses (graphing and drawing), to better measure higher-order mathematics and science skills (Cotter et al., 2020). At this time, the scoring of image response-based items necessitates human raters trained to score items based on comprehensive scoring guides. However, scoring errors and inconsistencies in scoring may arise due to rater effects such as severity or leniency, and other nuisance variables, such as fatigue or distractions, or simply due to differences in decision making based on human judgment (Zhang, 2013). One approach to mitigating human scoring errors is utilizing double rating designs where two different human raters score responses independently. However, this method increases workload, time, and costs. Another approach is to use automated scoring in place of a second human rater, which is low cost, consistent, and fast.

Large-scale assessments have utilized machine-supported scoring for some time (Attali & Burstein, 2006; Foltz et al., 1999; Sukkarieh et al., 2012; Yamamoto et al., 2017). However, these approaches have focused on automated scoring of text-based responses, and did not take more recent advances in using neural networks for deep learning of complex classification and prediction tasks into account. Artificial neural networks can also be used for text generation such as summarization, generation of questions and answers, as well as item stems in addition to automated scoring of existing items (von Davier, 2018, 2019). Furthermore, ANNs can be shown to perform estimations of parameters of complex probability functions that are both analytically intractable and difficult to estimate (Radev et al. 2019). Finally, ANNs can be shown to be universal function approximation methods (e.g., Hornik et. al. 1989; Zhou, 2020) and thus are ideally suited not only to generate text and approximate intractable estimation problems but also to mimic and even improve upon human scoring.

While machine learning can also be applied to non-text-based item responses, such as images (Pawlowicz & Downum, 2021; Yu et al., 2018), the automated scoring of constructed responses created using graphical input (e.g., images, graphs) has yet to be utilized in educational large-scale assessment. This study explores the automated scoring of image responses from a TIMSS 2019 item using ANNs, providing a comparison between feed-forward and convolutional approaches. The results have wider implications for ILSAs, showing how ANN-based scoring systems for image responses can be on par with human raters and utilized in a cost-efficient manner.

Background

To better assess students' higher-order cognitive processes, computer-based technology is needed to expand the range of tasks and response modes. Assessments that include a variety of computer-enhanced item types improve inferences about student cognition and narrow the gap between what can be seen and what can be measured (Zenisky & Sireci, 2002). Constructed-response items, in particular, are thought to allow measurement of a wider range of abilities and cognitive processes (Parshall et al., 2009). Additionally, they are thought to can capture higher-order thinking by asking students to draw shapes or pictures to demonstrate their understanding. In 2019, the Trends in International Mathematics and Science Study (TIMSS) used a variety of computer-based constructed-response items in their Problem Solving and Inquiry (PSI) tasks. PSIs are sets of items that simulate real-world situations, allowing students to apply process skills and knowledge across interrelated problems (Cotter et al., 2020). PSI items that require students to draw a diagram, create, modify, or annotate an image as a response aim at improving TIMSS' measurement of science and mathematics achievement. These response types provide researchers with information about student ability based on a performance that goes beyond entering a number or writing a short response.

While using image-based constructed-response items is beneficial to ILSAs, current scoring practices limit their widespread use. These items require human raters for scoring, which tends to be costly, time-consuming, and potentially less reliable than the scoring of choice-type responses such as those found in multiple choice items. . While scoring guides are designed to be comprehensive, they are not exhaustive; when students can draw responses freely, developing an inclusive scoring guide of all possible responses is unfeasible. Thus, some responses may require more complex decisions based upon human judgment, especially when those responses do not correspond with the existing examples in the scoring guide. When faced with these responses, raters may choose to treat them differently based on their interpretations of the guide and the response itself (Bejar, 2012). Differences in human judgment as well as effects arising from misinterpreted or only partially internalized scoring guides may lead to systematic scoring errors or inconsistencies in scoring. Additionally, rater effects such as severity and leniency, where raters may (or may not) be more apt to give students the benefit of the doubt than other raters, can impact scores' reliability and consistency (Zhang, 2013). In the context of international assessment, this can lead to systematic differences of how human-scored items are functioning across participating countries and limit comparability of the results obtained from such items (Oliveri & von Davier 2011; von Davier & von Davier, 2007; von Davier et al. 2018).

Machine-based scoring can mitigate effects of human scoring in assessments, either by serving as a second scorer or replacing human scorers entirely. Machine learning algorithms apply the same decision rules to all responses, making its scoring procedure objective and consistent (Williamson et al., 2012). Furthermore, automated scoring is fast and can classify thousands of responses in a few seconds (Zehner et al., 2016). ETS's e-rater system, for example, is used in the Graduate Record Examinations (GRE) for its essay portion. The essays are scored by a human rater and the e-rater system, with a second human rater used only when there is a vast disagreement between the two. Since its installment, the e-rater system has improved the GRE's scoring accuracy and validity (Zhang, 2013).

The Programme for International Student Assessment (PISA) also found similar benefits after implementing a machine-supported coding system. Unlike the e-rater system, PISA's coding system requires a perfect match approach between students' responses and responses from a data base (consisting of student responses from past PISA cycles that have been classified as correct or incorrect by human scorers). The machine automatically scores perfectly matching responses while responses with no perfect match are scored by a human rater (Yamamoto et al., 2017). While this approach showed to be very accurate for perfectly matching responses and reduced human workload, it was limited to short response items with high regularity in responses. Thus, this type of automated scoring could only be implemented for a fraction of the item pool (OECD, 2020).

Sukkarieh et al. (2012) utilized a more general approach for ETS' character-by-character highlighting item types. Using an algorithm from bioinformatics developed for gene sequence comparisons, the authors compared the highlighting of multilingual text sequences from students with responses provided by experts. As opposed to a perfect match approach, this system evaluated the highlighted elements of the text sequences as falling between the minimum or the maximum evidence of matching expert responses. Student responses that did not match the minimum or exceeded the maximum were given no credit. Ultimately, the algorithm improved scoring accuracy by eliminating misclassifications of responses due to irrelevant extra characters, or randomly missing characters, and focusing on similarity measures rather than exact matching.

New machine-based scoring approaches are more extensive and adaptable than the previously mentioned systems and other automated scoring procedures used currently in ILSAs. Machine learning has opened the door for more flexible scoring of responses for assessments. ETS' c-Rater system, for example, is a machine learning-based automated scoring system that makes classifications based on models trained to recognize key linguistic features unique to text-based items (Liu et al., 2014). Another example is the clustering system used by Zehner et al. (2016), where models were trained to identify semantic concepts in text-based items, cluster them, and apply scores to each cluster.

Machine learning can also be extended beyond text-based responses, namely to image responses, which has been illustrated in disciplines outside of large-scale assessments. Machine-based recognition of images in the medical field has been used for the last decade, and recent studies have shown that automated scoring of cell images to judge whether they are cancerous using algorithms is just as accurate as classifications made by trained doctors with years of experience (Rimm et al., 2018). Automated scoring of images has also been used to identify organ decay, where a deep learning algorithm was trained to recognize what stage of fibrosis livers were in based on postmortem biopsies (Yu et al., 2018).

Machine learning has even been used in the arts as well, where Lecoutre et al. (2017) automatically classified paintings by artistic style to index large artistic databases. Their approach involved training algorithms to identify essential elements of 25 different artistic styles, eventually producing a model that had 62% accuracy. Recently, deep learning was used to classify ancient pottery fragments (known as sherds) in an archaeology study conducted by Pawlowicz and Downum (2021). The authors were able to train a model to classify pictures of sherds from Arizona by decorative style with an accuracy comparable to, and sometimes even higher than, expert archaeologists they recruited.

Besides Rimm et al. (2018), the aforementioned studies utilized a specific type of machine-learning known as artificial neural networks (ANNs). ANNs were designed to mimic how neurons communicate in the human brain, enabling machines to learn and classify based upon a system of layers and nodes. The system's basic structure is an input layer, where the image information is received, hidden layers where the data is processed, and an output layer where classifications are made. Nodes function as interconnected processing elements within each layer, identifying pixels of images that certain classes have in common with one another. Neural network models can be trained on text sequences or on a set of images, where they learn to identify features unique to each class. Then, the models are tested on a set of validation data where they classify images they have never seen before based upon the iterations of training (O'Shea & Nash, 2015).

Two primary ANN approaches used for image classification are feedforward neural networks (FFNs) and convolutional neural networks (CNNs). FFNs are considered a form of traditional ANN with the basic three-part structure of input, hidden, and output layers. As their name suggests, the information processed by the nodes is only passed on in one (forward) direction. CNNs, on the other hand, have additional convolutional layers and pooling layers that occur before the fully connected layers (i.e., the FFN architecture). These additional layers allow the information to be processed in a non-linear manner and make CNNs especially adept at pattern recognition (O'Shea & Nash, 2015).

Despite the success of using ANNs to classify images in various fields, the procedure has yet to be implemented by any ILSAs. Thus, the current study was designed to assess both the automated scoring of image responses and to compare ANN approaches for potential use for image classification in large-scale assessments. The classification problem in this study is implemented as a supervised learning task based on an image response item from the TIMSS 2019 assessment that was administered as computer-based assessment in about half of the countries participating in TIMSS.

## Methods

*Item Selection and Rationale*

The study utilizes image responses from the third item in the TIMSS 2019 Building PSI task. This item asked students to draw a shed's back and side walls on a grid according to given specifications. Students received full credit for drawing the back of the shed and its sides correctly and partial credit if only the back wall was correct; all other responses were considered incorrect. This item was difficult for students—on average, only about 26% of students received full credit, and 11% received partial credit.

This item, in particular, was chosen because it would test whether automated scoring using neural networks could produce correct scoring results on par with human raters for complex TIMSS item response types. If the ANNs could score this complex item with high accuracy, it is expected to be just as successful for less complex items. Because the Building PSI item allows for partial credit and is scored with two points, it presents an additional challenge to the learning algorithm which has to distinguish between three score categories (0, 1, 2) instead of two (0 and 1 for incorrect and correct). Additionally, there is much variety in the responses within each score category, presenting an even greater challenge of associating images that may appear quite different with the same category. For example, full credit responses may show the back wall and sides of the shed in any orientation, as long as their measurements are correct. Incorrect responses present an even greater challenge, as they may be blank, off task, or may show shapes that are similar to the correct ones but differ in dimensions or angles.

*Response Variability*

Image responses that received full credit tended to be uniform in appearance, but the orientation of the shed's back and side walls varied between student responses. Additionally, the grid's border was considered a part of the image and students who used the border to compose one or more sides of the shape were still given credit if the dimensions were correct. Figure 1 includes some examples of full credit image responses.

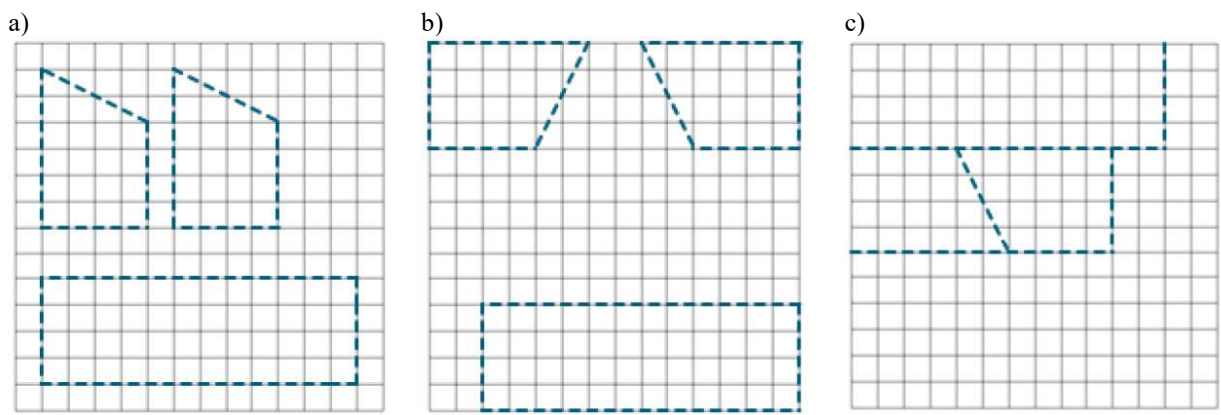

**Figure 1.** Examples of Full Credit Responses

Image responses given partial credit were even more diverse than those that received full credit. Partial credit responses tended to vary from containing one shape (the back wall only) to containing multiple shapes. Figure 2 includes some examples of partial credit image responses.

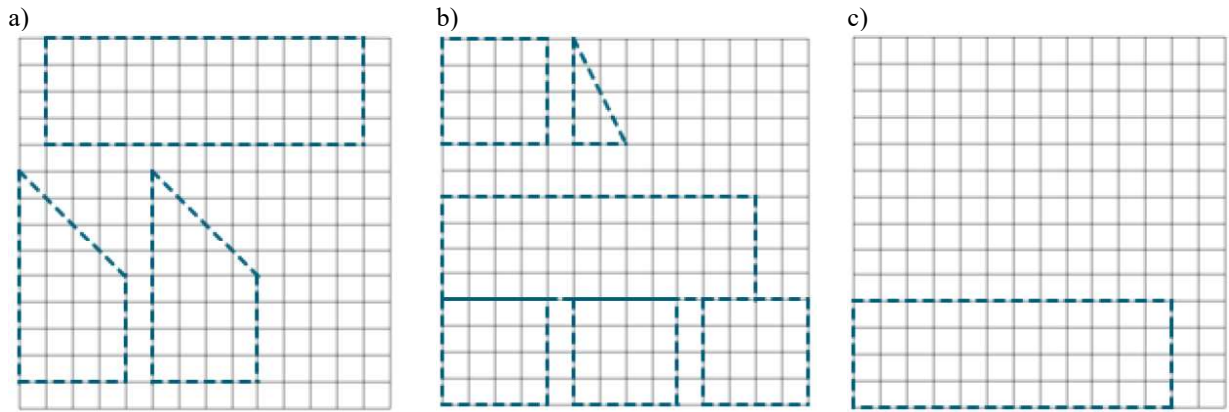

**Figure 2.** Examples of Partial Credit Responses

Finally, incorrect image responses tended to be blank (the grid was left empty), off task, or have inaccurate shapes. While blanks are easy to identify and score, off task responses were the most diverse, ranging from unintelligible scribbles to geometric art. On the other hand, images with inaccurate shapes were responses where students had attempted to answer the item but did not have the correct dimensions to receive full or partial credit. Figure 3 includes some examples of incorrect responses.

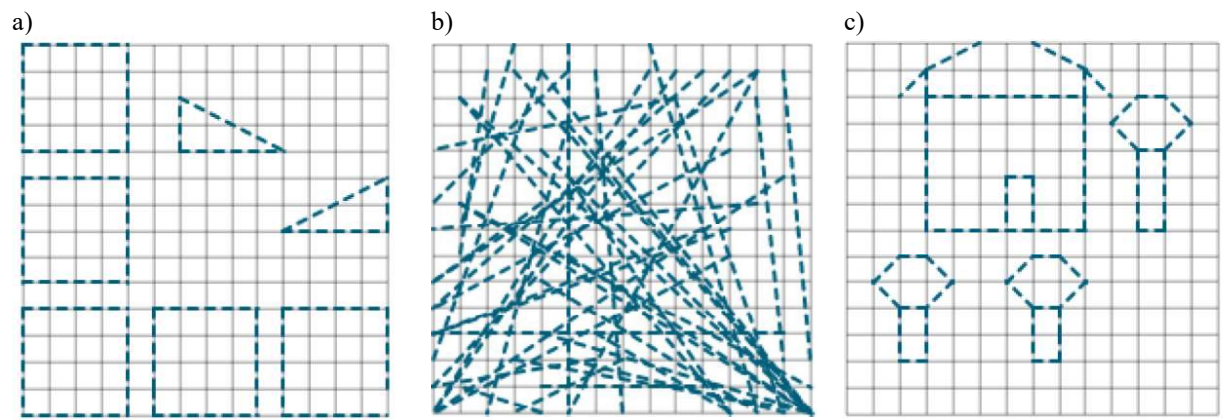

**Figure 3.** Examples of Incorrect Responses that have either Inaccurate Shapes (a) and are Off Task (b, c)

*Assignment and Incorrect and Inconsistent Response Removal*

Responses were collected from 23 countries that took the digital TIMSS 2019, as well as from three TIMSS 2019 benchmarking participants (e.g., regional entities, such as provinces and municipalities). Before pre-processing, the sample was split for simple hold-out validation where the image responses were assigned to a training and a validation sample. The assignment was done using stratified random sampling on a by-country basis. Within each country the responses were clustered by score category (incorrect, partial credit, full credit). Then, 70% of responses in each cluster were randomly assigned to the training sample. The remaining image responses were assigned to the validation sample. This data splitting method was chosen so that there was roughly equal representation of responses from each country and each score category in both datasets. For neural network training, it is desirable to have a larger training sample so that the models are exposed to more responses. However, if the training sample is too large, then the validation sample will be smaller and the results potentially not as generalizable (Chollet, 2018). A split of 70/30% was chosen to maximize the number of responses in the training sample while still having a large number of responses in the validation sample.

The image responses were also additionally assessed by an independent rater at the TIMSS & PIRLS International Study Center (ISC) to verify that the responses were given proper credit. This was done after analysis from a previous pilot study revealed that some image responses were scored incorrectly. More specifically, there were some scoring inconsistencies across the countries, particularly for image responses that included the shed's walls with the correct dimensions but also with extraneous lines. For example, image responses with the back wall of the shed split into three squares sometimes received no credit, or full or partial credit depending on the country. Inconsistencies such as these are likely due to some raters giving the benefit of the doubt to students and because the scoring guide did not contain information about whether to give credit when extraneous lines were present together with shapes that had the correct dimensions.

Initially, 416 image responses that were scored improperly by human raters (relative to the scoring guide) were removed from the study's modeling samples and put in a separate folder for later use. However, during this study's modeling process, examination of misclassified image responses revealed 9 additional incorrectly scored image responses from the validation sample. These responses were also set aside, and accuracies on the clean validation sample run again. Only these revised accuracies will be reported. Of the 425 removed responses, 193 were scored incorrectly, and 232 had common patterns of shapes that were scored inconsistently.

The study's final sample available for training and validation was composed of 14,737 image responses, with 10,238 images in the training sample and the remaining 4,499 images in the validation sample. Due to the difficult nature of the item, the majority of image responses were incorrect, with 63.66% of the training sample and 63.75% of the validation sample consisting of incorrect responses. 25.51% of the training sample and 25.54% of the validation sample received full credit. Partial credit image responses were the least prevalent, making up only 10.83% of the training sample and 10.71% of the validation sample.

*Pre-Processing and Array Creation*

After assignment, the images were converted from color to grayscale and their saturation was doubled to increase the contrast between the drawn shapes and the background grid. This process was done in R using the *magick* package (Ooms, 2021). The images were then saved as PNG files, separated by training and validation samples. Figure 4a shows an image response after pre-processing.

Next, the images were arranged into arrays that could be used for neural network modeling with the *keras* package in R (Allaire & Chollet, 2021). Using the *EBImage* package, the images were resized to be 64x64 pixels. 64 was chosen as the pixel size after preliminary testing found that a pixel size of 28 lost too much information while using larger pixel sizes took too much time for the models to run without improving classification rates noticeably. The images were then arranged into two arrays, one for training and one for validation (based upon their assignment). Figure 4b displays an example of an image response from an array using the *ggplot2* package. The corresponding image response classifications given by the human raters were also saved in an array using integer label format (rather than one-hot encoded format).

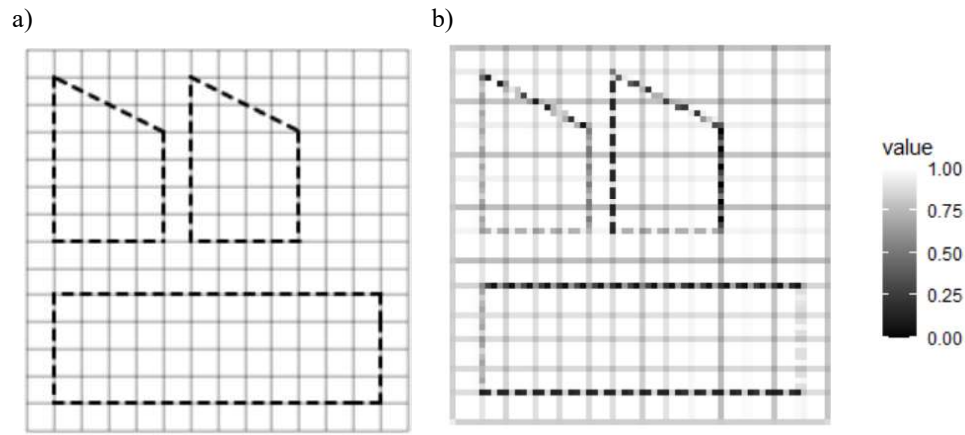

**Figure 4.** Examples of an Image Response After Pre-Processing (a) and in an Array (b)

*Training the Models*

All models were trained using the *keras* package in R (Allaire & Chollet, 2021) and compiled using the sparse_categorical_crossentropy loss option. All models had the input shape of 64x64x1 and had the final fully connected layer with three units, representing incorrect, partially correct, and fully correct. For every CNN model, the convolutional layers used a kernel size of 3x3 and were always followed by a 2-dimensional max pooling layer with a pool size of 2x2.

First, modeling was conducted using 25 and 50 epochs of training to determine how model accuracy and loss fared compared to the time it took to run the models. In this stage of modeling, all models were trained using the Adam optimizer (Kingma & Ba, 2014). Also, CNNs with zero padding and no padding were generated, as well as models with and without dropout layers. Models with dropout layers had dropout layers following each max pooling layer and fully connected layer, except for the final fully connected layer in the model. Each dropout layer used a dropout rate of 0.25, indicating that 25% of inputs would be dropped, at random, from the layer during each iteration of training.

After the ideal number of training epochs were determined, the Adam, AdaMax (Kingma & Ba, 2014), and Nadam (Dozat, 2016) optimizers were compared with zero and no padding, and either with or without dropout layers. The FFNs at these stages had two fully connected layers, with the first having 128 units. The CNNs at these stages had either one convolutional layer with 32 filters or two convolutional layers with 32 and 64 filters, respectively.

Finally, modeling turned to using different architectures with the best optimizer. Varying numbers of convolutional and fully connected layers were tested with different padding types. At this stage, all models included dropout layers. Each additional convolutional layer included 32 more filters than the previous layer. Each additional fully connected layer included 64 more units than the previous layer.

At each stage of the modeling process, model accuracy and loss were compared using the clean validation sample after training was complete.

Once the three stages were complete, the most accurate models were tested on the 425 incorrect and inconsistently scored image responses (which were not included in the clean validation sample). This was done to test how much the models disagreed with the human raters for these image responses and how many incorrectly scored image responses they could classify correctly (giving the correct amount of credit).

Results

*Testing Models on Clean Validation Data*

*Stage 1—Varying Number of Epochs*

The modeling process provided comparisons between FFNs and CNNs when trained under the same conditions. Additionally, a comparison was made between the CNNs and a highly accurate FFN model identified in a previous study at the ISC as the best option. This previous study used the same item, modeling various epochs, optimizers, and architectural structures of FFNs only. The most accurate model from that study was trained on 300 epochs and trained using the Nadam optimizer; it had 90.35% accuracy on the clean validation sample and a loss of 0.30.

In total, 50 models were created and trained. At each modeling stage, the CNNs outperformed the FFNs for accuracy in classifying the clean validation sample. The FFNs tended to be around 7 to 10% less accurate than the CNNs, and even the most accurate FFN model from the previous study at the ISC was, on average, 2% to 6% less accurate than most of the CNNs.

In the first stage of modeling, it was found that models with 50 epochs of training tended to have slightly higher accuracy than those with only 25 epochs of training. This was true for both the FFNs and CNNs, where models trained over 25 epochs had an accuracy of 80.77% for the FFN and accuracies of 92.82 to 94.75% for the CNNs while models trained over 50 epochs had an accuracy of 83.86% for the FFN and accuracies of 92.60% to 95.00% for the CNNs (see Table 1).

Based upon these results, it was decided that 50 epochs of training would be used for subsequent models.

It was also found that the CNNs with one convolutional layer instead of two tended to be less accurate and have higher loss. Thus, modeling with only one convolutional layer was not pursued further.

Table 1. Model Accuracies and Loss on Validation Data—25 epochs vs. 50 epochs

| Model | 25 epochs | | 50 epochs | |
| --- | --- | --- | --- | --- |
| | Loss | Accuracy | Loss | Accuracy |
| FFN | 0.5092 | 80.77% | 0.4133 | 83.86% |
| 1 convolutional layer CNN with no padding | 0.3576 | 93.02% | 0.4647 | 93.35% |
| 2 convolutional layer CNN with no padding | 0.2685 | 94.75% | 0.3200 | 94.11% |
| 2 convolutional layer CNN with no padding and dropout layers | 0.1815 | 93.80% | 0.2107 | 95.00% |
| 1 convolutional layer CNN with zero padding | 0.3218 | 92.82% | 0.4435 | 93.18% |
| 2 convolutional layer CNN with zero padding | 0.3472 | 93.44% | 0.3668 | 92.60% |
| 2 convolutional layer CNN with zero padding and dropout layers | 0.1781 | 94.33% | 0.2386 | 94.35% |

*Stage 2—Varying Optimizers*

For all three optimizers used in this study, the CNN models had higher accuracy and lower loss than the FFN model, with the accuracy difference ranging from around 9 to 13%. The Nadam optimizer yielded the highest accuracy of 95.38%, while the highest accuracy for models trained using Adam and AdaMax optimizers were 95.00% and 94.58%, respectively (see Table 2).

Table 2. Model Accuracies and Loss on Validation Data with Different Optimizers

| Model | Adam | | Nadam | | AdaMax | |
| --- | --- | --- | --- | --- | --- | --- |
| | Loss | Accuracy | Loss | Accuracy | Loss | Accuracy |
| FFN | 0.4133 | 83.86% | 0.3440 | 86.97% | 0.5367 | 78.42% |
| 2 convolutional layer CNN with no padding | 0.3200 | 94.11% | 0.4211 | 93.55% | 0.2910 | 94.29% |
| 2 convolutional layer CNN with no padding and dropout layers | 0.2107 | 95.00% | 0.2071 | 95.38% | 0.1649 | 94.58% |
| 2 convolutional layer CNN with zero padding | 0.3668 | 92.60% | 0.3429 | 94.58% | 0.3527 | 93.40% |
| 2 convolutional layer CNN with zero padding and dropout layers | 0.2386 | 94.35% | 0.2165 | 94.80% | 0.1729 | 94.40% |

On average, Nadam tended to provide the highest accuracy and had lower loss than models compiled using the Adam optimizer. While AdaMax yielded models with the lowest loss overall, the accuracies were slightly lower than the other two optimizers. It was decided that subsequent models would be trained using the Nadam optimizer.

It was also decided that subsequent CNN models would include dropout layers. This was done because the models with dropout layers tend to avoid overfitting while providing comparable accuracy. Dropout layers compensate for overfitting the training data by removing some of the neural pathways at each training epoch. This limits activations from becoming strongly correlated and thus reduces over-training. Consequently, when the model is applied to the validation data, it will have accuracy comparable to the training data. The reduction in overfitting is evidenced by the lower loss values for the models with dropout compared to their non-dropout counterparts. Moreover, for the Adam and Nadam optimizers, the models with dropout layers have slightly higher accuracies.

*Stage 3—Varying Layer Numbers*

In general, the more convolutional and fully connected layers present in a model, the higher the accuracy and lower the loss when tested on clean validation data. Consistent with the pattern observed thus far, the FFNs had lower accuracies than the CNNs. The highest accuracy achieved by an FFN with the Nadam optimizer trained on 50 epochs was 86.97% when that FFN had two fully connected layers, with the next highest accuracy being 83.80% when that FFN had three fully connected layers. The CNNs, on the other hand, had accuracies ranging from 93.55 to 97.71%. Results for the most accurate CNN models for each additional convolutional layer can be found in Table 3.

**Table 3.** Model Results for Validation Data with Different Number of Layers

| Number of Layers | | | | | |
|---|---|---|---|---|---|
| Convolutional | Fully Connected | Model | Loss | Accuracy | Number of misclassified responses |
| 0 | 2 | FFN from Previous Study* | 0.3020 | 90.35% | 434 |
| 5 | 3 | CNN with zero padding | 0.0939 | 97.71% | 103 |
| 4 | 4 | CNN with zero padding | 0.1013 | 97.24% | 124 |
| 3 | 4 | CNN with zero padding | 0.1097 | 96.58% | 154 |

*Trained on 300 epochs

The most accurate model was a CNN with five convolutional layers, two fully connected layers, and zero padding. This model had a loss of 0.09 and an accuracy of 97.71%, misclassifying only 103 of the 4,499 image responses in the clean validation sample. Classification accuracy was highest for full credit responses, classifying 99.04% correctly, and second highest for incorrect responses, classifying 97.94% correctly. The model classified 93.15% of partial credit responses correctly, which is one of the higher classification accuracies for this score category amongst all of the CNN models.

In comparison, the best FFN model created in the previous study was 7.36% less accurate and misclassified 434 of the 4,499 validation responses. It had lower classification accuracy for each score category than the CNN models, especially for partial credit image responses. This FFN model had the highest classification accuracy for incorrect image responses, classifying 94.74% correctly. It had a classification accuracy of 89.56% for full credit image responses. It had the lowest classification accuracy for partial credit image responses, only classifying 66.18% correctly.

Furthermore, the accuracy of image classifications by country varied less for CNN models compared to the FFN models. For the most accurate CNN model, the percent of correct classifications by country ranged from 92.91 to 99.28%. The average number of misclassified image responses across the countries was 4. For the best FFN model from the previous study, the percent of correct classifications by country ranged from 83.89 to 96.94%. The average number of misclassified image responses across the countries was 17.

*Comparing Misclassified Image Responses for FFNs and CNNs*

Image responses misclassified by the FFN models tended to have larger differences in scores between the human rating and the machine rating than the CNN models. Specifically, the FFN models would regularly give full credit to image responses that should have been classified as incorrect. The FNN models tended to associate more lines with more credit; thus, many of the image responses misclassified as receiving full credit were those that were off task or had inaccurate shapes with many lines. The FFN models also associated the angular lines of the sides of the shed (the roof) with full credit. As a result, many image responses where the sides of the shed were drawn correctly, but the back wall was not were given full credit when they should have received none (see Figure 5). The FFN models also exhibited misclassification of images that should receive partial credit—the algorithms tended to misclassify partial credit responses that contained only a rectangle (with no other shapes in the grid) as incorrect.

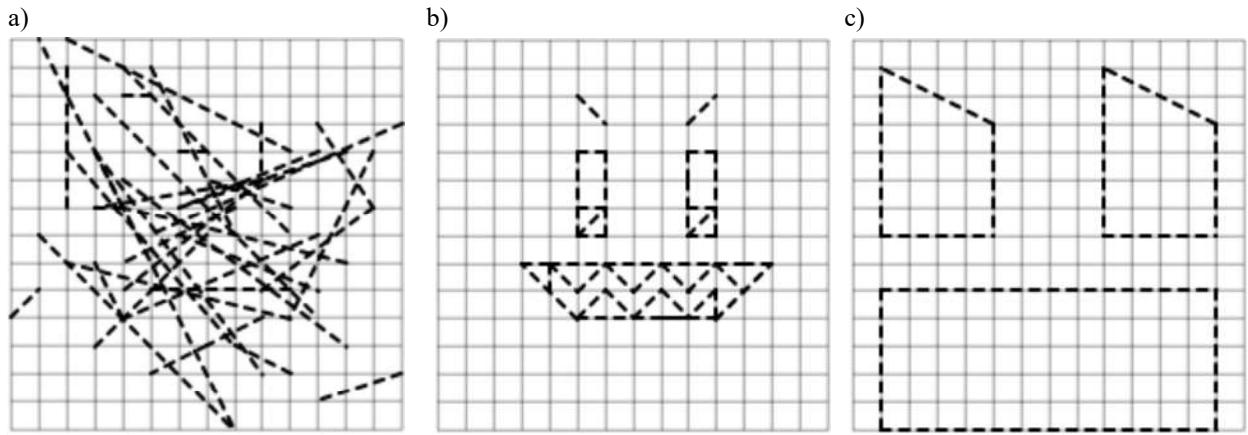

**Figure 5.** Examples of Incorrect Image Responses Misclassified as Full Credit by FFN Models

The CNN models tended to misclassify fewer images overall. And if misclassification occurred, the image responses were put into one of the adjacent score categories, with misclassifications of incorrect as full credit rarely occurring. Additionally, the CNN models did not associate more lines with full credit; thus, off task responses that were obviously incorrect to the human eye (such as examples a and b in Figure 1) were not given any credit. Like the FFN models, the CNN models also associated correct diagonal lines of the roof or correct sides of the shed with full credit (even if the back wall was incorrect). They also tend to classify images where the sides of the roof were drawn close together as only deserving partial credit or no credit at all when they should have received full credit (see Figure 6).

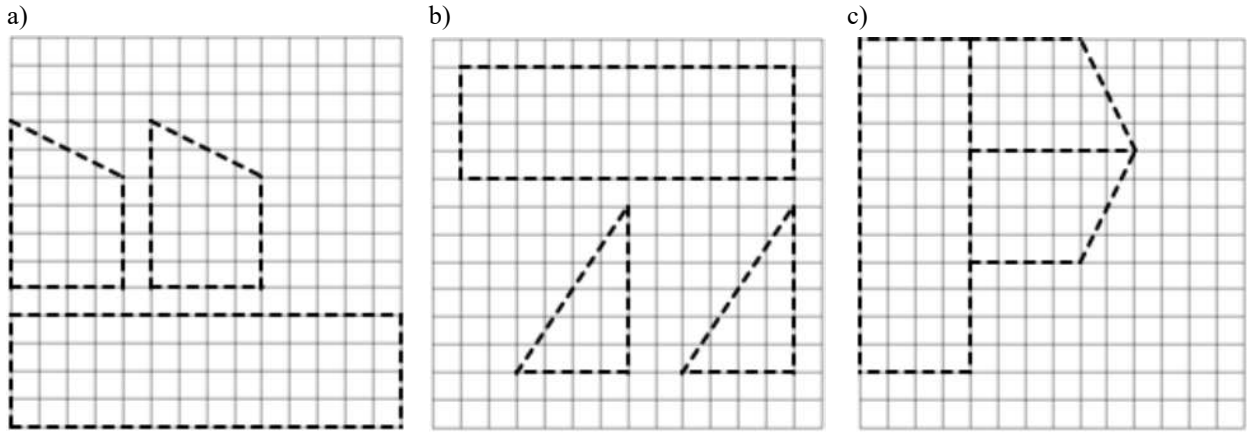

**Figure 6.** Examples (rare) of Incorrect (a) and Partial Credit (b) Responses Misclassified as Full Credit and a Full Credit (c) Response Misclassified as Partial Credit by the CNN models.

*Testing Models on Incorrectly or Inconsistently Human-Scored Image Responses*

Overall model accuracy decreased, as expected, when the most accurate models were applied to the validation set containing incorrectly scored and inconsistently scored image responses. For the 425 incorrect and inconsistently scored responses in the sample, the CNN models disagreed with the (incorrect or inconsistent) human ratings for 76.71 to 80.71% of them. The disagreement was lower for the FFN model, with only 69.41% of image responses being scored differently. For incorrectly scored image responses only, the CNN models classified 87.05% to 89.64% correctly (scores matching an independent rater at the ISC). In comparison, the most accurate FFN from the previous study only classified 79.79% of them correctly.

The most accurate CNN model classified correctly 167 of the 193 (86.53%) that human raters scored incorrectly, which was the highest accuracy for incorrectly scored responses of the CNN modes (see Table 6).

**Table 4.** Model Performance for Inconsistent and Incorrect Image Responses

| Number of Layers | | | Overall Disagreement | | Correct Classifications for Incorrect Responses | |
|---|---|---|---|---|---|---|
| Convolutional | Fully Connected | Model | Number | Percent | Number | Percent |
| 0 | 2 | FFN from Previous Study* | 295 | 69.41% | 131 | 67.88% |
| 5 | 2 | CNN with zero padding | 343 | 80.71% | 167 | 86.53% |
| 4 | 4 | CNN with zero padding | 335 | 78.82% | 163 | 84.46% |
| 3 | 4 | CNN with no padding | 326 | 76.71% | 157 | 81.35% |

*Trained on 300 epochs

*Testing Models on Model Based Expected vs. Human-Scored Responses*

As discussed above and shown by example, some human scorers produce classifications that are either incorrect (do not follow the scoring guides) or inconsistent (the same response could be scored differently by different scorers). Due to this observation it can be questioned whether human generated scores are indeed the gold standard that represents the best training target. Instead of using an expert generated second scoring process, the assessment data available from other tasks can be utilized to generate the proverbial 'second opinion'. Each student responds to more items than the building task and these are used to produce data on their overall achievement. The latter can be modeled using item response theory (IRT; Lord & Novick, 1968) and population modeling to generate what is called plausible (proficiency) values. Population or conditioning models take achievement data (i.e., information from the IRT scaling) and reports of student academic context variables (i.e., from student questionnaires) into account to estimate more accurate proficiency values (i.e., plausible values) than can be estimated with IRT alone (see, for example, Mislevy & Sheehan, 1990; Thomas, 1993; von Davier et al, 2006; von Davier & Sinharay, 2013; von Davier et al. 2018).

This study uses human rater generated scores and IRT-based expected or maximally likely scores and compares them to select images for the training data. These IRT-based estimates can be produced by utilizing the general student performance obtained from the TIMSS 2019 population modeling (von Davier, 2020) and the reported item parameters from the TIMSS 2019 IRT scaling (Foy et al., 2020) as fixed values in an IRT model to generate an estimate of the probability that a student with proficiency $\theta$ solves an item with parameters $a, b, c$ (for example). More precisely, we have a human rater generated score $x_{in[r]}$ that assigns no credit, partial credit, or full credit to the encounter of a student $n$ with task $i$ based on rater $r$'s evaluation of the image. Then, we can also generate expected scores based on the IRT item function that assigns response probabilities to each potential score level of the task, either using expected a priori (EAP) estimation or maximum a priori (MAP) estimation,

$$\text{EAP} = y_{in[exp]} = round\left[\sum_{x=0}^{2} xP(X = x|\theta, a, b, c)\right]$$

or

$$\text{MAP} = y_{in[max]} = max_{(x=0..2)}\{P(X = x|\theta, a, b, c)\}$$

where EAP estimation is the weighted average of all response's probabilities for a student $n$ and MAP estimation is the highest response probability for student $n$. This approach allows for comparing the human rater generated score $x_{in[r]}$ and the IRT-based expected or maximally likely score $y_{in[exp]}$ and $y_{in[max]}$, and to use the agreement between the two to select images for the training data.

A training sample of images based on the agreement between the two provides a likely much 'cleaner' base for training. While human raters are providing the correct ratings in most cases, having an additional expected rating—based on student achievement data that is highly related—allows us to select those responses that are most likely correct or incorrect, or receive partial credit, based on both human rating and most likely performance predicted using all available information on the student.

An estimate of student ability was used to predict the expected response on the graphical response item. The achievement items used to estimate ability are a mix of both machine scored multiple-choice items and human-scored constructed response items. The choice of ability estimate is secondary here, so we opted for the posterior mean of ability based on the population modelused operationally in TIMSS 2019, which utilizes both item responses and contextual variables in the latent regression to reduce bias in secondary analyses (von Davier, 2020). The IRT item parameters were for the graphical response item were $a = 0.62, b = 0.93, c_1 = -0.88, c_2 = 0.88$, and can be found in the [TIMSS 2019 International Database](#) (Fishbein, et al., 2021). For ANN modeling, only the clean sample of responses was used, meaning that all incorrect and inconsistently scored image responses were excluded.

After computing the response probabilities and the likely IRT-based scores (MAP and EAP), it was revealed that the MAP-based scored matched 72.74% of the human-based scores. However, this method did not produce any expected scores of partial credit (one-point-responses). This is because the likelihood of receiving partial credit on the item under the IRT formula is never higher than the likelihood of receiving no credit or full credit. Therefore, any training sample constructed would be devoid of partial credit responses, and thus the model would not learn to distinguish them from incorrect or full credit.

The EAP method, on the other hand, produced expected scores that matched 55.83% of the human-based scores. Within these expected scores, there is an overrepresentation of partial credit responses, with 42.27% of the EAP-based scores consisting of partial credit compared to only about 10% being present in the human-based sample. As a consequence, there was a severe underrepresentation of expected full credit responses, with only 6.43% of the EAP-based scores consisting of full credit compared to 25% present in the human-based sample.

As a result of the distributions of both the expected and maximally likely scores, two methods of analyses were chosen. The first method utilized the MAP-based scores, but removed all responses given partial credit by the human raters. This filtering increased the percentage of matching responses between the MAP-based scores and the human-based scores to 81.54%. The second method utilized an averaging of the MAP-based and EAP-based scores, which enabled analysis using a training sample that included partially correct responses, but also had a higher number of full credit responses than with EAP alone. 62.98% of the average MAP and EAP-based scores matched the human-based scores, with 25.49% of the expected scores consisting of partial credit and 23.20% consisting of full credit.

Examination of the average response probabilities revealed that the IRT model predicts much higher probabilities within respective score categories when the IRT-based scores and the human scores match. Conversely, nonmatching scores have probability distributions that are closer together. Additionally, the IRT model tends to overpredict incorrect responses because the item was difficult for students (see Tables 5 and 6).

Table 5. Average posterior probabilities for MAP-based scores

| Sample | P(incorrect) | P(partial credit) | P(full credit) |
| --- | --- | --- | --- |
| Matching—incorrect | 78.50% | 10.04% | 11.46% |
| Matching—full credit | 23.14% | 14.12% | 62.75% |
| No match—incorrect | 29.22% | 15.43% | 55.35% |
| No match—full credit | 59.90% | 14.63% | 25.47% |
| Overall | 64.02% | 11.61% | 24.37% |

Table 6. Average posterior probabilities for average MAP and EAP-based scores

| Sample | P(incorrect) | P(partial credit) | P(full credit) |
| --- | --- | --- | --- |
| Matching—incorrect | 85.26% | 8.35% | 6.39% |
| Matching—partial credit | 55.39% | 15.62% | 28.98% |
| Matching—full credit | 23.14% | 14.12% | 62.75% |
| No match—incorrect | 48.66% | 15.47% | 35.86% |
| No match—partial credit | 57.85% | 12.44% | 29.71% |
| No match—full credit | 59.90% | 14.63% | 25.47% |
| Overall | 63.26% | 11.82% | 24.92% |

The next step in the process involved assigning the response to samples for neural network modeling. For both the MAP and EAP approaches, 70% of the responses where there was a match between the IRT-based score $y_{in[max]}$ (MAP) or $\frac{y_{in[max]}+y_{in[exp]}}{2}$ (average MAP and EAP) and the human-based score $x_{in[r]}$ were randomly assigned to the training sample ($T_{match}$), and the remaining 30% were assigned to a validation subset sample ($V_{match}$). All responses where there was no match between the scores were assigned to a separate validation sample ($V_{no\_match}$). Additionally, random samples were also constructed for comparison. These randomly generated training samples ($T_{rand}$) had the exact same number of responses (by score category) as the IRT-based training samples, but were randomly assigned regardless of whether the IRT-based scores and the human scores matched. The remaining responses were assigned to one comparison validation sample ($V_{rand}$).

For neural network modeling, FFN and CNN models underwent 50 epochs of training with the Nadam optimizer and the predictions were compared with the human-based responses. The CNN models were all constructed with zero padding and included dropout layers. For the MAP approach (with no partial credit responses), 7,489 responses were assigned to the training samples ($T_{match}$ and $T_{rand}$), 2,427 responses assigned to the mismatch validation sample ($V_{no\_match}$), and 3,221 responses assigned to the match validation subset ($V_{match}$); 5,648 responses were assigned to the random comparison validation sample ($V_{rand}$). For the average MAP and EAP approach, 6,461 responses were assigned to the training samples ($T_{match}$ and $T_{rand}$), 5,455 responses assigned to the mismatch validation sample ($V_{no\_match}$), and 2,821 responses assigned to the match validation subset ($V_{match}$); 8,276 responses were assigned to the random comparison validation sample ($V_{rand}$). See Table 7).

Table 7. Sample Sizes for ANN Modeling with IRT-based Scores and Random Comparison Samples

|  | IRT-based Samples | | | Random Comparison Samples | |
|---|---|---|---|---|---|
|  | Training ($T_{match}$) | Mismatch validation ($V_{no\_match}$) | Match validation subset ($V_{match}$) | Training ($T_{rand}$) | Validation ($V_{rand}$) |
| MAP-based Scores | 7,489 | 2,427 | 3,221 | 7,489 | 5,648 |
| Average MAP and EAP-based Scores | 6,461 | 5,455 | 2,821 | 5,455 | 8,276 |

For both IRT-based approaches, the CNN models outperformed the FFN models for every validation sample. Additionally, performance was higher for the validation sample where the IRT and human scores matched than for the sample where they did not match. For the MAP-based scores the difference was only around 1%, but was around 4% for the average MAP and EAP-based scores. Moreover, compared to the random sample results, the validation sample where the scores did not match performed worse, but the validation samples where the scores matched tended to have slightly higher accuracy. For the CNN models, the random sample validation results were higher than the mismatch validation sample by about 1% for MAP-based scores and 3% for the averaged MAP and EAP-based scores. The differences between the match validation subset and the random comparison validation sample were much smaller, with only a 0.01 to 0.03% improvement for the MAP-based score models and a 0.5 to 1.5% improvement for the average MAP and EAP-based score models. In general, the models utilizing MAP-based scores had higher performance than the average MAP and EAP-based scores, with CNNs reaching 98.63% accuracy compared to 96.70% (see Tables 8 and 9). The higher performance for this approach is likely due to the absence of partial credit responses, which were challenging to classify for every model trained throughout this study.

Table 8. Model Performance for MAP-based Scores

| Number of Layers | | | Mismatch validation sample | | Match validation subset sample | | Random comparison validation sample | |
|---|---|---|---|---|---|---|---|---|
| Convolutional | Fully Connected | Model Type | loss | accuracy | loss | accuracy | loss | accuracy |
| 0 | 2 | FFN | 0.3297 | 86.86% | 0.1662 | 93.70% | 0.2075 | 91.41% |
| 5 | 2 | CNN | 0.1293 | 97.90% | 0.0859 | 98.54% | 0.0714 | 98.51% |
| 4 | 4 | CNN | 0.1155 | 97.36% | 0.0676 | 98.63% | 0.0682 | 98.62% |

Table 9. Model Performance for Average MAP and EAP-based Scores

|  Number of Layers | | | Mismatch validation sample | | Match validation subset sample | | Random comparison validation sample | |
| --- | --- | --- | --- | --- | --- | --- | --- | --- |
| Convolutional | Fully Connected | Model Type | loss | accuracy | loss | accuracy | loss | accuracy |
| 0 | 2 | FFN | 0.6827 | 72.41% | 0.3163 | 88.90% | 0.5402 | 78.78% |
| 5 | 2 | CNN | 0.3535 | 92.80% | 0.155 | 96.70% | 0.1958 | 95.86% |
| 4 | 4 | CNN | 0.4391 | 91.20% | 0.1781 | 96.56% | 0.2368 | 94.80% |

Discussion

In each stage of the modeling process, the CNN models outperformed the FFN models both in terms of accuracy and loss. Not only did the CNN models reach accuracies upwards of 98%, but they also had very little overfitting when dropout layers were utilized. While both the CNN and FFN models had the lowest classification accuracy for partial credit image responses, the CNN models still had around 90% accuracy compared to the 60% accuracy for the FFN models. These results indicate that despite being a three-point item, the CNN models were able to automatically score most responses correctly.

The benefit of using an ANN-based automated scoring system is that trained models can classify thousands of images in mere seconds with extremely high accuracy and little cost. ILSAs could train neural network models for image response-based items in place of second human raters. Any differences in classifications between the human raters and the automated scoring models could be reviewed by an expert rater at the ISC. The expert rater would only need to classify a fraction of the responses that a full-time rater would.

Moreover, if CNNs are used instead of FFNs, the pool of image responses to review would be even smaller. For example, the number of image responses in the study that would need to be evaluated by an additional expert human rater was only 103 for the best CNN model instead of 434 for the best FFN model. This means that if the CNN model were used instead of the FFN model, the expert rater would have to evaluate only 2.29% of the validation image responses instead of 9.65%.

Furthermore, the neural network models revealed inconsistencies and errors in the human ratings. As previously mentioned, early models correctly classified some image responses that were incorrectly classified by the human raters. Even after an independent rater at the ISC reviewed the entire sample of image responses, some incorrectly scored responses remained. These responses were classified correctly by the CNN models but appeared as misclassifications. While these results exemplify human rater effects, they also evidence just how accurate neural networks can be.

To utilize CNNs operationally for automated scoring in ILSAs, models could be trained on a subset of image responses from past cycles or on responses from field test data collection. If response sample sizes are small, images can be transformed (e.g., rotated, flipped, and cropped) to increase the data's training sample size. Model accuracy would be tested on the remaining sample of validation responses. The most accurate model would then be used for automated scoring after the main data collection is complete.

An extra measure of quality control can be added to the scoring by constructing training and validation samples based on IRT expected scores. The results indicate that, in general, models trained on data where IRT-based and human scores matched performed similarly to models trained with randomly generated data samples (including matching and not matching scores) when applied to all data not included in the training samples (the validation samples). However, if we look closer at the results for the random comparison validation samples compared to the samples with only matching responses (the match validation subsets), we observe that the IRT expected score models had slightly higher accuracies and lower loss. This means that while the IRT-based ANN models have high classification in their own right, they work especially well for responses where there is agreement between human-based and IRT-based scores.

This IRT-based approach to model construction could be used as a pre-modeling quality check, where IRT-based ANN models could be produced prior to actual modeling with randomly composed samples to help identify incorrectly scored image responses. This could be accomplished by examining image responses where there was a mismatch between predicted ANN scores and human-based scores, particularly for image responses where the IRT-based scores and human-based scores are the same. Utilizing this approach may aid an independent rater during the data cleaning process.

Despite the overwhelming benefits of automated scoring using neural networks, two potential limitations exist. First, the process of training CNN models can be time-consuming, especially if using a computer's central processing unit (CPU) rather than graphics processing units (GPU). That being said, the longest time it took to train a CNN model in the study using CPU was roughly two hours. Additionally, once a model is trained, it can be applied to new data in seconds, regardless of the processing unit. The training can also be done in advance using data collected prior to operational use, and in parallel on clusters of CPUs.

The other potential limitation of using automated scoring with CNNs is that the high quality training data is needed, and training data must be cleaned before use. It is recommended that training data be carefully assessed so that incorrect image responses are re-scored and inconsistencies in ratings are identified. This process can be time-consuming for independent raters, but accurate training data is essential to training accurate models. The use of model based expected responses will be further explored for this purpose. One approach would be to clean the training data utilizing expected responses generated using a variety of models to impute or predict the most likely response category based on the other responses, but also utilizing covariates from context questionnaires.